\documentclass[10pt,twocolumn,letterpaper]{article}

\usepackage[pagenumbers]{cvpr} 

\definecolor{cvprblue}{rgb}{0.21,0.49,0.74}
\usepackage[pagebackref,breaklinks,colorlinks,allcolors=cvprblue]{hyperref}

\usepackage{comment}
\usepackage{adjustbox}
\usepackage[whole]{bxcjkjatype}

\DeclareMathOperator*{\argmin}{arg\, min\, }

\renewcommand{\vec}[1]{\boldsymbol{#1}}

\def\rm#1{\mathrm{#1}}

\def\paperID{4280} 
\def\confName{CVPR}
\def\confYear{2026}

\title{Towards Imperceptible Watermarking Via Environment Illumination for Consumer Cameras}

\author{%
Hodaka Kawachi$^{\dagger}$ \quad
Tomoya Nakamura$^{\dagger}$ \quad
Hiroaki Santo$^{\dagger}$ \quad
SaiKiran Kumar Tedla$^{\ddagger}$\\
Trevor D. Canham$^{\ddagger}$ \quad
Yasushi Yagi$^{\dagger}$ \quad
Michael S.~Brown$^{\ddagger}$\\[2pt]
$^{\dagger}$The University of Osaka \quad
$^{\ddagger}$York University
}

\begin{document}
\maketitle
\begin{abstract}
This paper introduces a method for using LED-based environmental lighting to produce visually imperceptible watermarks for consumer cameras. Our approach optimizes an LED light source's spectral profile to be minimally visible to the human eye while remaining highly detectable by typical consumer cameras. The method jointly considers the human visual system's sensitivity to visible spectra, modern consumer camera sensors' spectral sensitivity, and narrowband LEDs' ability to generate broadband spectra perceived as ``white light'' (specifically, D65 illumination). To ensure imperceptibility, we employ spectral modulation rather than intensity modulation. Unlike conventional visible light communication, our approach enables watermark extraction at standard low frame rates (30--60 fps). While the information transfer rate is modest---embedding 128 bits within a 10-second video clip---this capacity is sufficient for essential metadata supporting privacy protection and content verification.
\end{abstract}
    
\section{Introduction}
\begin{figure}[tpb]
    \centering
    \includegraphics[width=1.0\linewidth]{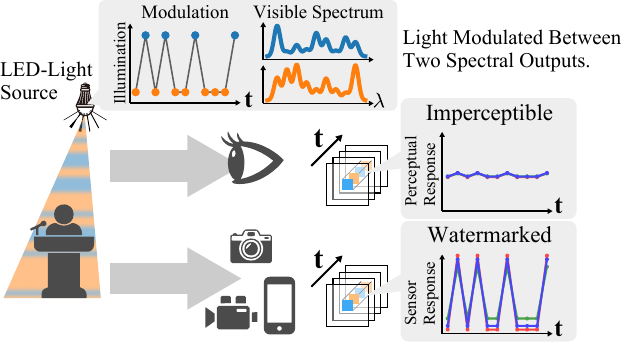}
    \caption{Overview of the proposed method. Using two optimized spectra as lighting and switching them at 15 fps, the flickering is completely imperceptible to humans but detectable by the camera, acting as a watermark.}
    \label{fig:overview}
    \vspace{-0.5em}
\end{figure}    

This paper explores a method for using LED-based environmental lighting to physically ``watermark'' a scene in a way that is invisible to viewers and unnoticeable in recorded video. This watermark can help prevent unauthorized recording when integrated with camera manufacturers, as cameras can be designed to detect the watermark and disable recording in restricted areas. This approach also provides a means to verify video authenticity, revealing when a scene was not captured at the intended location. This technique holds promise as a future privacy measure and a safeguard against deepfake content.

To achieve this, we propose an active illumination approach using narrowband LEDs capable of high-speed modulation. To reliably embed and extract information with consumer cameras operating at 30–60 fps, illumination must be modulated at or below half the camera's frame rate (i.e., 15 fps), following Shannon's sampling theorem. However, modulation within this range is highly perceptible to humans and can cause discomfort, including risks such as photosensitive epilepsy. This constraint makes direct temporal modulation impractical for non-intrusive applications.

To overcome this limitation, we employ temporal spectral modulation, leveraging differences in spectral sensitivity between CMOS sensors and the human visual system~\cite{jiang2013space}. Instead of modulating light intensity, we adjust the LED spectra to meet three key criteria: (1) minimal detectability by human observers, (2) maximum detectability by CMOS sensors, and (3) a white appearance equivalent to standard D65 illumination. By exploiting this spectral sensitivity disparity, we generate optical changes imperceptible to humans yet detectable by cameras, enabling low-frequency spectral modulation without visual discomfort while preserving information embedding.

Although the system transmits data at a relatively low rate of 15 bits per second, it can embed up to 128 bits within a 10-second video clip. This capacity is sufficient to encode essential metadata, such as location information, recording permissions, and timestamps--key details that support privacy protection and content verification.

\vspace{0.5em}
\noindent{\bf Contributions} To the best of our knowledge, we present the first method for optically embedding information into LED illumination sources using spectral modulation, enabling extraction from video recordings of the scene. By leveraging an optimization approach that accounts for differences in spectral sensitivity between humans and cameras, we achieve spectral variations detectable only by cameras. We validate our method's feasibility through a user study with a prototype lighting system and various consumer cameras, demonstrating detection from videos captured across different devices and through unknown in-camera processing, while confirming that the signal remains below the human visibility threshold.
Beyond watermarking, our approach may serve as a human-invisible, camera-centric active cue for broader vision tasks---for example, near-infrared ``dark-flash'' illumination that aids geometry/appearance estimation~\cite{darkflash_iccv21}---with related explorations indicating similar potential for depth sensing and for hyperspectral (and HS+3D) reconstruction using commodity RGB cameras~\cite{activestereonet_eccv18,park_iccv07,shin_cvpr24,shin_cvpr25}.
\section{Related Work}
\label{sec:related_works}
This section reviews related work in visible light communication (VLC), spatial watermarking, hyperspectral lighting, and the characteristics of human and camera color perception. 

\vspace{0.5em} \noindent{\bf Visible Light Communication (VLC).}
VLC is a wireless communication technology that transmits data using visible light, typically emitted by LEDs~\cite{tanaka2000wireless, ieee2011ieee, karunatilaka2015led, tsonev2014light, sevincer2013lightnets, khalighi2014survey}.
Unlike radio-frequency (RF) communication, VLC exploits the rapid on-off switching capabilities of LEDs, allowing them to function as both illumination and data-transmitting sources.
This high-frequency flickering, imperceptible to the human eye, makes VLC promising for applications such as indoor positioning and secure data transfer in RF-sensitive environments (e.g., hospitals and airplanes).

To achieve VLC with image sensors, the low frame rate of typical cameras necessitates data transmission at relatively low frequencies near the human perceptual threshold (30–60 Hz). Several studies have explored rolling shutter image sensors, which can capture high-frequency flickering signals due to the sequential exposure of pixels across the sensor~\cite{Liang, Chow2015, Ji}.
While this approach enables high-frequency VLC with cameras, it is impractical for one-to-many communication because variations in shutter speed and exposure time require calibration or prior knowledge of each camera's settings.
Thus, to ensure universal compatibility, it is crucial to use low-frequency, imperceptible lighting that remains undetectable to the human eye yet reliably detectable by camera sensors.

\vspace{0.5em} \noindent{\bf Spatial Embedding.}
Recent CVPR/ICCV work has rapidly advanced invisible digital watermarking for validating generative content~\cite{zhang_cvpr25_omniguard,bui_iccv25_trustmark}. Similarly, in the spatial domain, information can be embedded by projecting patterns using structured light or projection mapping~\cite{Amiri, Geng_11}. Although these spatial methods can embed more information through spatial addressing and multiplexing, they usually require pre-mapping the scene and precise alignment between the projector and surfaces, which makes them difficult to deploy and maintain in large or changing environments. In addition, making projected cues truly imperceptible in everyday settings remains challenging.

\begin{figure}[htpb]
    \centering
    \includegraphics[width=1.0\linewidth]{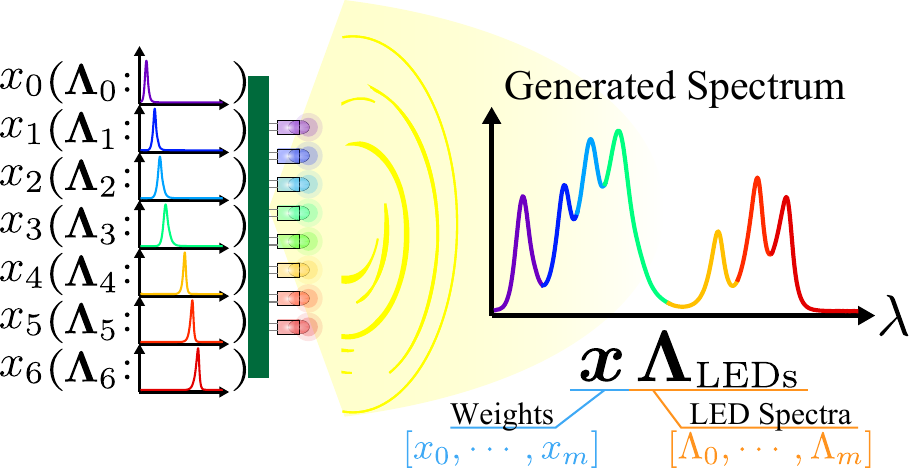}
    \caption{Configuration of a hyperspectral lighting system using narrowband LEDs. By controlling the emission intensity of each LED, a custom spectral output is generated through the weighted sum of individual LED spectra.}
    \label{fig:led_spectrum}
\end{figure}

\vspace{0.5em} \noindent{\bf Hyperspectral Illumination.}
Recent advances in narrowband LED technology have enabled the construction of hyperspectral lighting systems by combining multiple LEDs~\cite{zhao2020narrow}.
As shown in Fig.~\ref{fig:led_spectrum}, the spectrum of a narrowband LED lighting system can be customized by adjusting the emission intensities of individual LEDs, which combine as a weighted sum of each LED's spectral profile.
Traditionally, controllable hyperspectral lighting required specialized light sources, typically involving diffraction gratings or spatial light modulators~\cite{zheng2015separating}.
In contrast, the multi-LED approach offers a simpler and more accessible way to design lighting with tailored spectral characteristics, enabling customized spectral profiles.

\vspace{0.5em} \noindent{\bf Human Color Perception and Metamerism.}
Human color perception is governed by the trichromatic theory, which states that color vision relies on three types of photoreceptor cells in the retina, each sensitive to different regions of the visible spectrum. These sensitivities, often represented by the CIE XYZ color matching functions~\cite{cie1931proceedings, ws82}, serve as the foundation for standard color spaces that model human vision~\cite{agoston2013color, H84}. However, metamerism allows different spectral power distributions (SPDs) to produce identical color sensations for human observers~\cite{fairchild2013color, SSBR07}. This occurs because the brain interprets color based on the combined responses of the three photoreceptor types, meaning distinct light spectra can elicit the same perceptual response.

\begin{figure*}[t]
    \centering
    \includegraphics[width=1\linewidth]{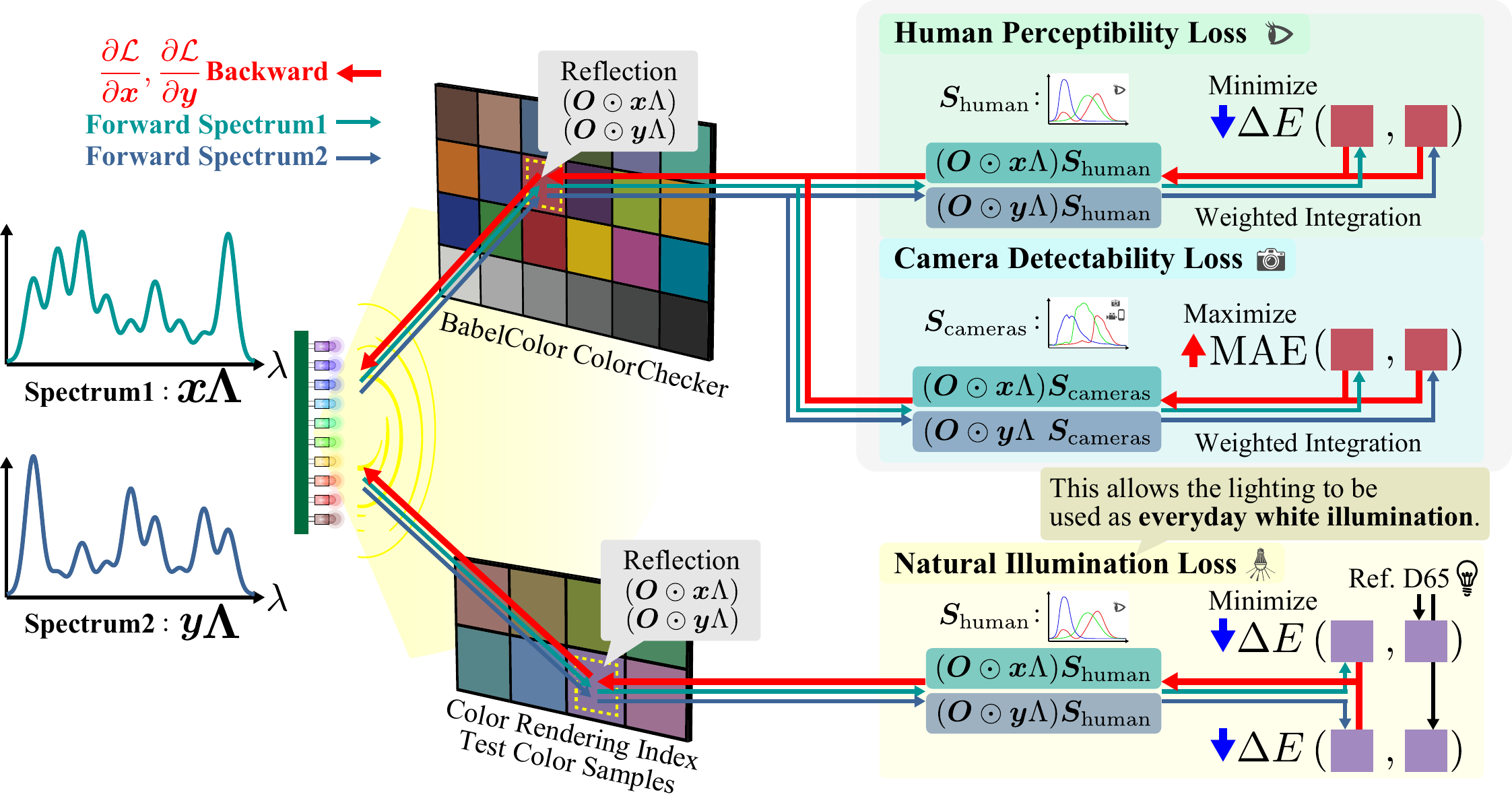}
    \caption{Proposed optimization pipeline. The objective functions ensure human imperceptibility, camera detectability, and natural white light for everyday use. These functions guide the optimization of LED intensity variables to achieve the desired lighting characteristics. Dark blue and dark green arrows indicate the forward pass for the two illuminations, while red arrows denote the backpropagation path used to optimize the LED intensities.}
    \label{fig:proposed_pipeline}
\end{figure*}

\vspace{0.5em} \noindent{\bf Comparing Camera and Human Spectral Sensitivity.}
Consumer cameras approximate human color perception using three primary color filters. However, achieving an exact match with human visual sensitivity is challenging due to technical and design constraints. Factors such as filter material properties, manufacturing costs, and brand-specific design choices create variations in camera spectral sensitivity. Due to differences in spectral sensitivities, cameras apply color space mappings to convert sensor-specific RGB values to a human-perceptual color space ~\cite{HLR01,hakki_color,kim2025ccmnet}. Linear transformations are commonly used to align the camera's color space with the human XYZ space, but these adjustments are inherently limited. As a result, camera sensors still differ from human vision in their spectral response~\cite{jiang2013space}.

\section{Proposed Method}
As discussed in Sec.~\ref {sec:related_works}, notable sensitivity differences exist between human vision and cameras. Leveraging these differences, we propose a method that optimizes spectral lighting to embed information detectable by cameras while remaining imperceptible to humans. This section details the spectral optimization process, the flicker detection technique for information retrieval, and the impact of noise on detection accuracy.

\subsection{Spectral Optimization}
We optimize two ambient light conditions that are imperceptible to humans but detectable by a camera to achieve communication through lighting. 
We denote these optimized ambient lights as $\vec{L}_1$ and $\vec{L}_2$. Switching between these lights at a rate equal to or lower than the frame rate establishes communication between the lighting system and the cameras. 
The two ambient light conditions are optimized to satisfy the following three requirements jointly:
\begin{enumerate}
\item The difference between the lights is imperceptible to humans.
\item The difference is detectable by the cameras.
\item The ambient light provides a natural color rendering suitable for everyday use.
\end{enumerate}
Let the reflectance characteristics of the objects be denoted as $\vec{R}_\rm{o} \in \mathbf{R}^{N\times l}$, where $N$ is the number of objects. 
The reflectance of the $i$-th object is denoted as $\vec{R}_\rm{o}[i] \in \mathbf{R}^{l}$.
Fig.~\ref{fig:proposed_pipeline} illustrates the optimization pipeline incorporating these requirements.

\vspace{0.5em} \noindent{\bf Human Perceptibility.}
The perceptibility of color differences to humans has been extensively studied in color difference research and is quantified using DeltaE2000~\cite{brainard2003color, stokes1992}. For our method, we use DeltaE2000($\Delta_\rm{E}$) to ensure that the reflected light from all objects remains imperceptible to humans. To meet this criterion, we define the following evaluation function:
\begin{equation}  
    \mathcal{L}_\rm{h} = \Delta_\rm{E}\left( \vec{s}_\rm{h}\left(\vec{R}_\rm{o}\odot\vec{L}_1\right), \vec{s}_\rm{h}\left(\vec{R}_\rm{o}\odot\vec{L}_2\right) \right),
\end{equation}  
where $\vec{s}_\rm{h}$ denotes the spectral sensitivity of the human visual system, $(\vec{R}_\rm{o} \odot \vec{L}_i)$ represents the element-wise product of the object reflectance $\vec{R}_\rm{o}$ and the light spectrum $\vec{L}_i$, and $\vec{s}_\rm{h}(\vec{R}_\rm{o}\odot\vec{L}_i) \in \mathbf{R}^3$ represents the observed XYZ values.

A difference of $\Delta_\rm{E} \leq 2$ is generally imperceptible to untrained observers when colors are viewed side by side~\cite{brainard2003color, stokes1992}. However, in practical applications, the human eye is more sensitive to low-frequency flicker, making it more likely to detect subtle differences when colors alternate over time, even if these differences are imperceptible in a static, side-by-side comparison.  Therefore, the evaluation function is minimized under this constraint, provided that the requirements of the other loss functions are satisfied.

\vspace{0.5em} \noindent{\bf Camera Detectability.}  
To ensure the watermark remains imperceptible, it is crucial that the flicker does not produce excessive pixel differences that could be easily noticeable to viewers. If the flicker in the captured video is too prominent, it may significantly degrade visual quality, making the video unsuitable for typical viewing. The following evaluation function is designed to ensure adequate detectability across all spectral channels while controlling pixel differences for multiple camera models:
\begin{align}  
    \mathcal{L}_\rm{c} &= -\min_{i} f\left( \rm{MAE}\left(\vec{v}_1[i], \vec{v}_2[i]\right) \right)\\  
    \vec{v}_1[i] &= \vec{s}_\rm{c}[i]\left(\vec{R}_\rm{o}\odot\vec{L}_1\right)\\  
    \vec{v}_2[i] &= \vec{s}_\rm{c}[i]\left(\vec{R}_\rm{o}\odot\vec{L}_2\right)\\  
    f(x) &= \begin{cases}  
        x & x < \tau_\rm{c}\\  
        \tau_\rm{c} & \text{otherwise}  
    \end{cases}  
\end{align}  
where $\vec{s}_\rm{c}[i]$ denotes the spectral sensitivity of the $i$-th channel for each camera, $\rm{MAE}$ is the mean absolute error, and $\tau_\rm{c}$ is a threshold. The function $f(x)$ clips the loss to a maximum of $\tau_\rm{c}$.

To align this objective with other losses in the optimization, which are formulated as minimization tasks, a negative sign is applied to  $\mathcal{L}_{\rm c}$. This converts it from a maximization objective to a minimization one, ensuring consistency across all loss terms.

The spectral sensitivity differences between humans and cameras vary across channels, and minimizing the average error across all channels can result in disproportionately high or low errors in specific ones. This approach prioritizes the channel with minimum error to prevent the embedding from being affected by an object's color or reflectance characteristics.

This formulation ensures that the flicker effect in the camera remains within a manageable range, keeping the average difference across all channels near the threshold $\tau_\rm{c}$. We set 
$\tau_\rm{c}$ to $1/256$, representing the detectable value in the context of 8-bit video quantization. This threshold allows for reliable detection, as the average includes colors with low brightness and those predominantly influenced by specific color channels, meaning not all pixel differences are strictly limited to $1/256$. Additionally, the loss function is averaged across multiple cameras to ensure broader applicability rather than optimizing for a single device.

\vspace{0.5em} \noindent{\bf Natural Illumination.}
Another criterion for this method is that the light must maintain a neutral white appearance with minimal spectral skew, ensuring suitability for everyday use.
Color rendering is conventionally evaluated using the Color Rendering Index (CRI), which measures color fidelity based on standardized samples.
CRI values range from 0 to 100, with higher values indicating more natural color rendering. Common household LEDs and bulbs typically have CRI values around 60, which is generally sufficient for everyday applications.

Our approach targets the standard D65 illumination (i.e., spectra that appear as daylight) and enforces a CRI of at least 60 per the CRI standard (eight test samples).
We define \(\mathrm{CRI}=100-4.6\,\Delta E\), where \(\Delta E\) is the average color difference to D65. To ensure \(\mathrm{CRI}\ge 60\), we minimize a hinge penalty on \(\Delta E\) (Eq.~\ref{eq:cri}).
\begin{align}
\label{eq:cri}
\mathcal{L}_{\mathrm{w}} &= \sum_{i=1}^8 \operatorname{ReLU}\!\left(\Delta_{1,i}-\tau_{\mathrm{w}}\right)
                         + \operatorname{ReLU}\!\left(\Delta_{2,i}-\tau_{\mathrm{w}}\right) \\
\Delta_{1,i} &= \Delta E\!\big( \vec{s}_{\mathrm{h}}(\vec{R}_{\mathrm{o}}[i]\odot \vec{L}_{\mathrm{ref}}),~
                                 \vec{s}_{\mathrm{h}}(\vec{R}_{\mathrm{o}}[i]\odot \vec{L}_{1}) \big) \\
\Delta_{2,i} &= \Delta E\!\big( \vec{s}_{\mathrm{h}}(\vec{R}_{\mathrm{o}}[i]\odot \vec{L}_{\mathrm{ref}}),~
                                 \vec{s}_{\mathrm{h}}(\vec{R}_{\mathrm{o}}[i]\odot \vec{L}_{2}) \big),
\end{align}
where \(\vec{L}_{\mathrm{ref}}\) is the D65 reference spectrum and \(\tau_{\mathrm{w}}=40/4.6\) enforces \(\mathrm{CRI}\ge 60\).

\vspace{0.5em} \noindent{\bf Weighted Optimization.}
The optimal two spectra are designed by adjusting and minimizing these three losses as follows:
\begin{equation}
    \vec{L}_1, \vec{L}_2 = \argmin_{\vec{L}_1, \vec{L}_2} \left(
        \vec{w}_\rm{h} \mathcal{L}_{\rm h} + \vec{w}_\rm{c} \mathcal{L}_{\rm c} + \vec{w}_\rm{w} \mathcal{L}_\rm{w}
    \right),
\end{equation}
where $\vec{w}_\rm{h}, \vec{w}_\rm{c}, \vec{w}_\rm{w}$ are the weights for each loss function, which determine the optimization direction.
These weights essentially involve a trade-off, so their selection is critical. We set the weights such that the color rendering loss becomes zero and the camera loss is approximately $1/256$. 
The human visual loss is minimized as much as possible while satisfying these conditions.

\vspace{0.5em} \noindent{\bf Reparameterization for Hyperspectral Illumination.}
As shown in Fig.~\ref{fig:led_spectrum}, hyperspectral illumination can be reproduced as a weighted sum of multiple LEDs. 
Therefore, rather than directly optimizing $\vec{L}_1$ and $\vec{L}_2$, we optimize the intensity of each LED. 
Let each LED's spectral profile be denoted as $\vec{\Lambda}i$ and their respective intensities be $x_i$ and $y_i$. 
The spectral compositions of the lighting follow:
\begin{equation}
\vec{L}_1 = \sum_{i=1}^{m} x_i \vec{\Lambda}_i,
\quad\vec{L}_2 = \sum_{i=1}^{m} y_i \vec{\Lambda}_i,
\end{equation}
where $m$ is the number of LEDs.
In this manner, reparameterization shifts the optimization target to $\vec{x}=[x_0,\cdots, x_m]$ and $\vec{y}=[y_0,\cdots, y_m]$.

%

\subsection{Detection of Flicker}
We now outline the detection algorithm. The supplementary material provides a more detailed explanation.

\label{sec:detection_flicker}

\vspace{0.5em} \noindent{\bf Frames per Second (fps) and Bits per Second (bps).}
In our setup, lighting fluctuations are encoded as binary values, with frame-to-frame pixel intensity changes representing bits (1 for a change, 0 for no change). This differential encoding generates a bit sequence based on frame differences across the image. According to Shannon's theorem, a recording rate of $n$~fps ideally enables communication at $n$~bps. However, real cameras have finite exposure times, which can introduce inaccuracies if phase shifts occur between illumination changes and the camera's sampling phase. To mitigate this, we set the recording rate to $2n$~fps, ensuring synchronization and achieving $n$~bps.

\vspace{0.5em} \noindent{\bf Signal in Dark Regions.}
In dark regions, 8-bit quantization can erase inter-frame differences, as pixel fluctuations are often too small to detect accurately.
To improve detectability, we set the camera detectability loss so that the average pixel fluctuation is approximately 1.
Pixels with values consistently below 128 across all channels and time are masked to prevent quantization loss.


\vspace{0.5em} \noindent{\bf Normalization.}
To ensure detection accuracy despite differences in scene brightness due to camera settings like exposure time and white balance, we normalize pixel intensity values across all channels, setting the mean intensity to 1. This normalization standardizes variations, enabling consistent results across different camera configurations.

\vspace{0.5em} \noindent{\bf Temporal Differential Calculation.}
With normalized data, we compute differences along the time axis to capture frame-to-frame changes representing encoded binary values. This is achieved by subtracting each frame from the previous one. At this stage, differential information is obtained at $2\times$sampling rate to mitigate phase shifts. For final processing, we return to the signal's original sampling rate by downsampling, selecting the phase that best captures the lighting fluctuations to address potential misalignment.

\vspace{0.5em} \noindent{\bf Signal Binarization.}
Finally, we compute the average brightness changes across the image for each time frame to capture fluctuations over time. To extract a binary signal that adapts to the scene's characteristics, we use an adaptive thresholding approach. We begin with a static threshold to estimate the average brightness fluctuation level when significant changes occur. This serves as a baseline for assessing typical fluctuation intensity. Based on this baseline, we then define a dynamic threshold set to half of the average fluctuation, ensuring consistent identification of significant brightness changes as binary signals.

\subsection{Noise Effects}
\label{sec:noise}
The effect of noise during image capture is an important factor in determining the ability to detect flicker.
As this algorithm averages the pixel differences across the entire image, the effect of noise is minimized.
Assuming the noise is additive white Gaussian noise with a mean of $0$ and variance $\sigma^2$, the mean $\tilde{\mu}$ and variance $\tilde{\sigma}^2$ of the noise relative to the overall image difference can be expressed as follows:
\begin{equation}
\tilde{\mu} = 0, \quad \tilde{\sigma} = \sigma\sqrt{\frac{2}{N}}
\end{equation}
Considering that recent video resolutions are around 1 megapixel, the effect of noise is sufficiently negligible.
\begin{figure*}[t]
    \centering
    \includegraphics[width=1\linewidth]{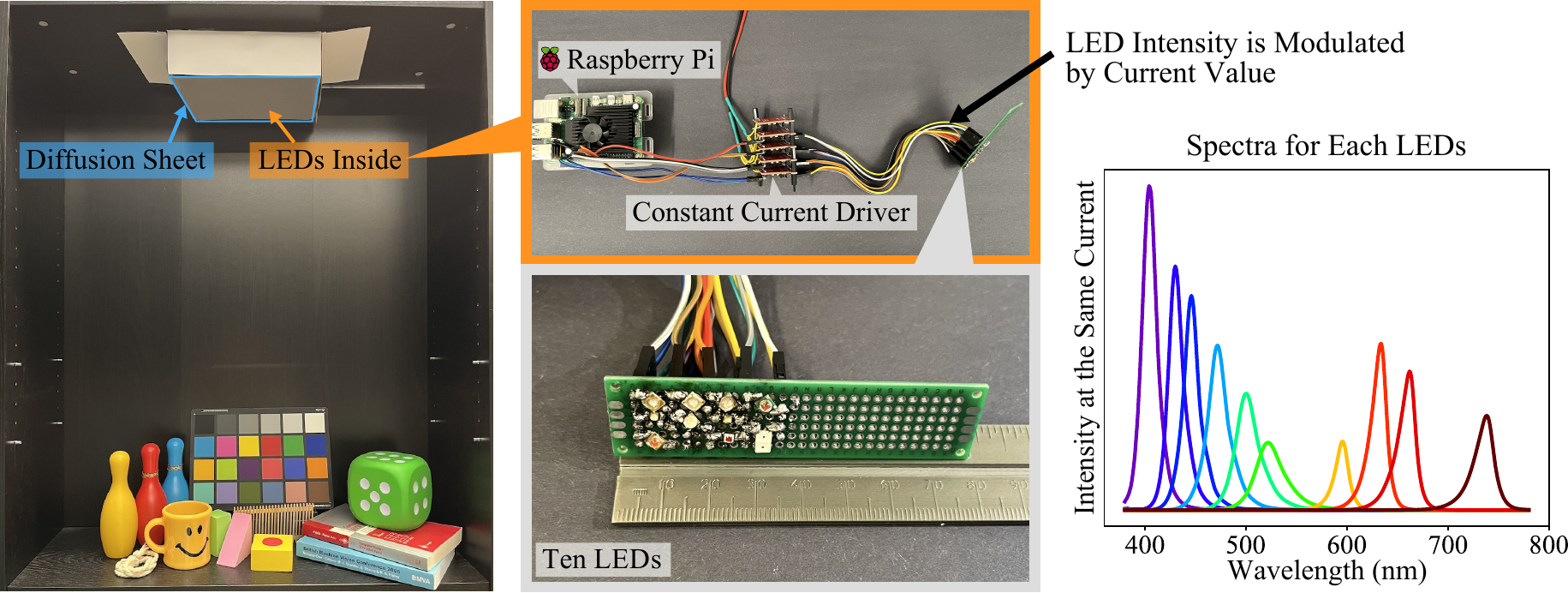}
     \caption{Lightbox using the prototype hyperspectral lighting. Each LED's intensity is adjusted using a constant-current chip and a Raspberry Pi. The spectral distribution of each LED is measured inside the light box by capturing the white patch of the color chart.}
    \label{fig:experiment_setup}
    \vspace{-0.5em}
\end{figure*}

\begin{figure}[t]
\centering
\includegraphics[width=1.0\linewidth]{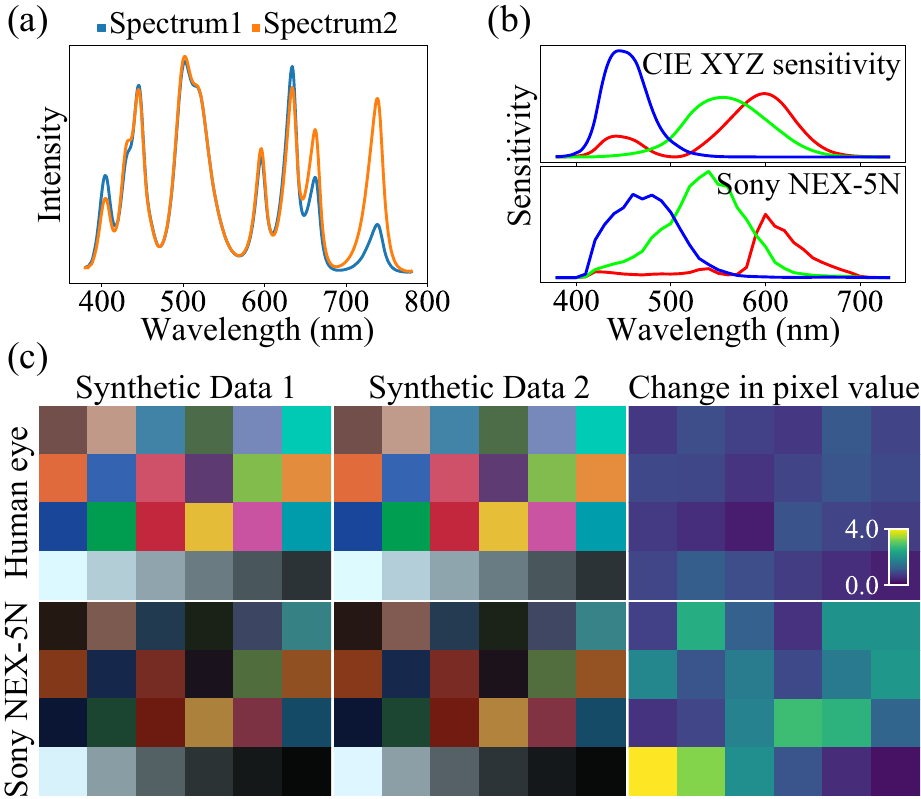}
\caption{
  (a) Plot of the two optimized spectra. 
  (b) The CIE 1931 Standard Observer XYZ Color Matching Functions and an example from the camera spectral sensitivity dataset~(Sony NEX-5N).
  (c) Synthetic rendered color checker and pixel value variations obtained using the optimized spectra.
}
\label{fig:spectral_optimization}
\vspace{-0.5em}
\end{figure}

\section{Experimental Results}
This section describes the experimental setup and evaluations used to validate the proposed method. First, we detail the implementation of spectral optimization, ensuring embedded information is detectable by cameras while remaining imperceptible to human observers. Next, we assess human sensitivity to the optimized flicker to confirm its imperceptibility. Finally, we quantitatively evaluate the system's communication accuracy across various environmental conditions and camera setups, examining its robustness and reliability in practical scenarios.

\subsection{Spectral Optimization}
\label{sec:exp_spectral}
We use a BabelColor ColorChecker chart with known reflectance properties to optimize the spectra. This chart consists of 24 color patches, each with a defined reflectance value. Spectral optimization is performed by treating the chart as the target object.

For the prototype lighting setup, we use ten types of LEDs with different wavelengths, as shown in Fig.~\ref{fig:experiment_setup}. LED brightness is modulated by adjusting the current through a constant-current chip using pulse-width modulation via a Raspberry Pi. During optimization, the duty cycle of each LED serves as the adjustable variable for brightness control.

Camera sensor characteristics for the optimization are taken from a publicly available dataset of spectral sensitivities for 28 cameras \cite{jiang2013space}. 
The optimization problem is solved using the Adam optimizer~\cite{kingma2014adam} with a learning rate of $0.01$, over 5000 iterations.
The hyperparameters for optimization, denoted as $\vec{w}_h, \vec{w}_c, \vec{w}_w$, are determined through grid search.
The hyperparameters used in this case are $\vec{w}_h = 0.15, \vec{w}_c = 0.05, \vec{w}_w = 0.8$.

The final spectra obtained through this optimization process are shown in Fig.~\ref{fig:spectral_optimization}.
The optimized pair yields a mean $\Delta E=0.2839$, i.e., imperceptible.

Based on the cameras' spectral sensitivities from the dataset, the average MAE across all channels was $1.843$ in 8-bit representation. Using the proposed method's definition of $\mathcal{L}_{\rm c}$, the minimum MAE across channels is $0.972$, sufficient to produce detectable pixel value differences even after 8-bit quantization.

\subsection{Human Sensitivity Evaluation}
\label{sec:exp_real_scene}
\noindent{\bf Experimental Setup.}
To assess the perceptibility of the proposed method, we evaluated human color perception under LED illumination optimized using the proposed spectra. For this evaluation, we used the lightbox setup shown in Fig.~\ref{fig:experiment_setup}, which contained various colored objects for participants to observe.

We prepared two flickering illumination patterns:
\begin{enumerate}
\item \textbf{Optimized Spectra:} Alternating illumination between the two LED-based spectra optimized by our method.
\item \textbf{Baseline Intensity Pattern:} A conventional VLC scheme that varies light intensity for data transmission (ratio 1:0.97).
\end{enumerate}
To equate the sensor-level signal, the baseline ratio was tuned such that, under the same camera and exposure, the mean per-frame pixel-value difference between its two intensity levels matched that observed between the optimized spectra.

Participants took part in the experiment under the following sequence of conditions:
\begin{enumerate}
  \item \textbf{Viewing Steady-State Baseline:} Participants first observed the objects under a steady optimized light for 10~sec to understand the arrangement of objects.
  \item \textbf{Randomized Observation of Flicker Patterns:} Participants then observed a randomly selected lighting pattern from four possible conditions:
  \begin{itemize}
      \item Optimized Steady State
      \item Optimized Flicker State
      \item Baseline Steady White Light
      \item Baseline Flicker White Light (intensity ratio of 1:0.97)
  \end{itemize}

  \item \textbf{Feedback on Perceived Flicker:} Participants were asked whether they noticed any flicker in the lighting after each observation.

  \item \textbf{Repeated Randomized Testing:} This process was repeated so that each lighting pattern was presented exactly five times to each participant, ensuring a balanced distribution across all conditions.
\end{enumerate}

\begin{figure}[t]
    \centering
    \includegraphics[width=1.0\linewidth]{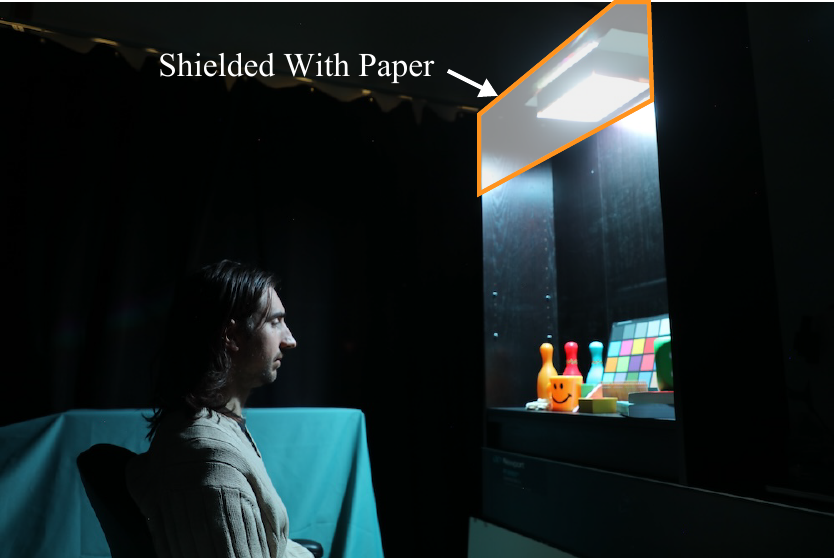}
    \caption{
    Our experimental setup shows a participant observing the optimized lighting patterns. Shielding prevents direct light exposure, allowing participants to observe the flicker indirectly through reflected light only.
    }
    \label{fig:human_eval}
    \vspace{-0.5em}
\end{figure}

\begin{table}[htpb]
    \centering
      \setlength{\tabcolsep}{3pt}        
      \renewcommand{\arraystretch}{0.98} 
      \begin{tabular*}{\linewidth}{@{\extracolsep{\fill}} lcc @{}}
    \toprule
        Illumination Type & Avg. detections (range) & Detection rate\\
        \midrule
        Optimized Steady  & 0.133 (0--2) & 2.67\% \\
        Optimized Flicker & 0.467 (0--2) & 9.33\% \\
        Baseline Steady      & 0.267 (0--2) & 5.33\% \\
        Baseline Flicker  & 4.933 (4--5) & 98.67\% \\
        \bottomrule
    \end{tabular*}
    \caption{Detection results. For each condition, 15 participants each performed five trials. ``Avg. detections'' is the average number of detections per participant over five trials (parentheses show the participant-wise min--max). ``Detection rate (\%)'' equals (avg. detections / 5) $\times$ 100.}
  \label{tab:detection_results}
\end{table}

\vspace{0.5em}
\noindent{\bf Results.}
We tested 15 participants (Table~\ref{tab:detection_results}). Each participant was exposed to four illumination conditions: (i) Optimized Steady State; (ii) Optimized Flicker State ($\Delta E = 0.2839$); (iii) Baseline Steady White Light; and (iv) Baseline Flicker White Light with an intensity ratio of 1:0.97 ($\Delta E = 0.6080$). Although $\Delta E$ values in this range are typically regarded as below the threshold of noticeability for static color differences, in our low-frequency flicker experiment, the Baseline Flicker White Light condition (1:0.97) was detected by nearly all participants (98.67\%), indicating that such small chromatic differences become clearly visible when presented as low-frequency temporal flicker.

By contrast, the optimized spectral modulation yielded a mean detection rate of only 9.33\%; moreover, no participant reported it in more than half of the five trials, indicating that it was effectively imperceptible for most observers.

For the static controls, false-positive rates were lower still (2.67\% for static optimized and 5.33\% for static white), likely reflecting minor fluctuations or momentary adaptation. Taken together, these results support spectral optimization as a viable strategy for low-frequency modulation in imperceptible watermarking applications. At the same time, the findings underscore that conventional intensity modulation is unsuitable in this regime due to human flicker sensitivity.

\begin{figure*}[t]
  \centering
  \includegraphics[width=1.0\linewidth]{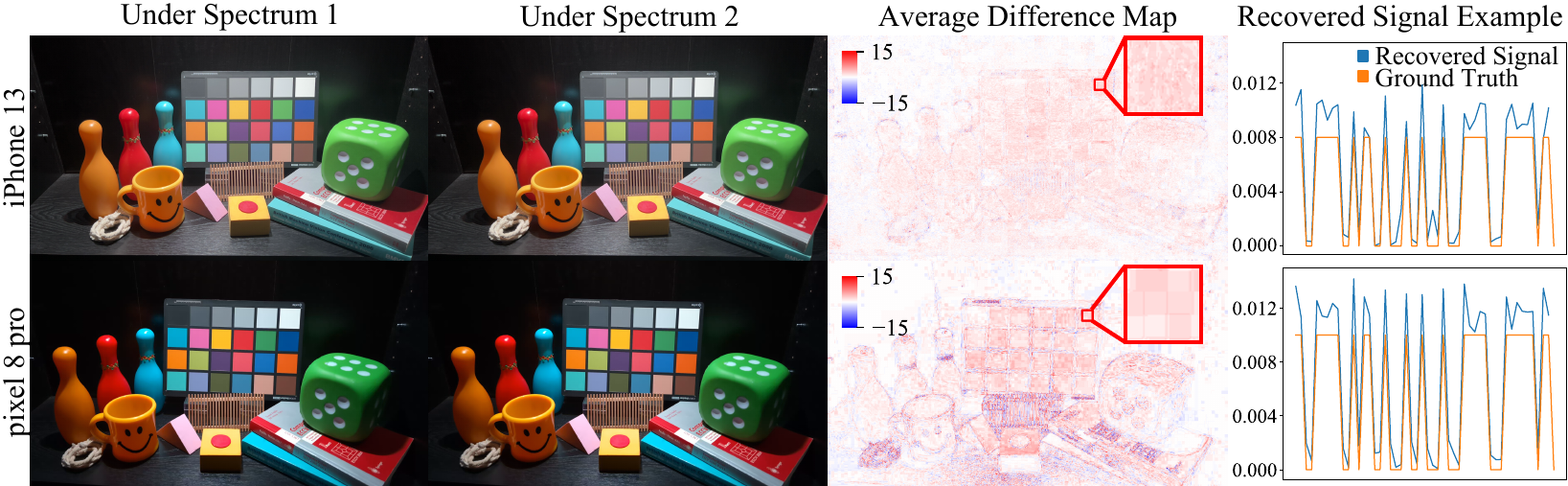}
  \caption{
    This figure illustrates the decoding results from a scene captured using two spectral lighting conditions, Spectrum 1 and Spectrum 2. The first two images display frames from the observed video---(left)~Spectrum 1 and (right)~Spectrum 2.  The third image displays the average RGB differences between these frames. The fourth plot shows the recovered signal values before binarization.
  }
  \label{fig:real_scene}
\end{figure*}
\begin{figure}[t]
  \centering
  \includegraphics[width=1.0\linewidth]{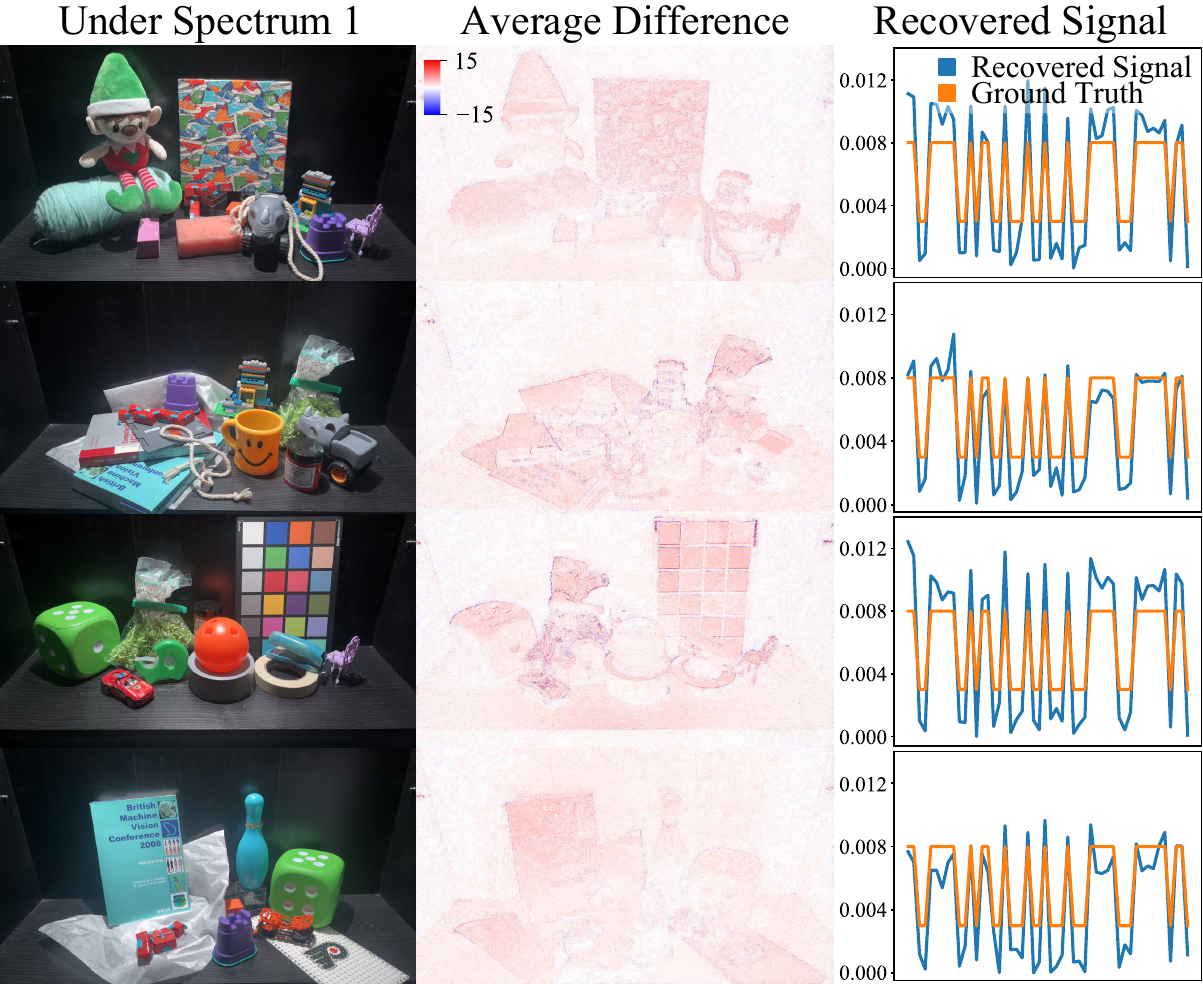}
  \caption{
    Experimental scenes were captured with iPhone 13 and evaluated, with objects of varying reflectance properties arranged as randomly as possible to assess robustness across lighting and reflectance conditions. The graphs in the figure display the recovered signal values before binarization.
    }
  \label{fig:multi_scenes}
\end{figure}

\subsection{Evaluation of Decoding}
\noindent{\bf Communication Accuracy in a Single Scene.}
First, we evaluated the communication accuracy of the proposed method in a controlled environment using a single scene. We conducted long-duration recordings with three cameras (iPhone 13, Pixel 8 Pro, and Nikon D7200), each receiving 6400 bits, over several minutes per device. This setup established a baseline measure of communication accuracy for each camera and assessed whether the proposed method could achieve consistent performance across different devices. The results are shown in Fig.~\ref{fig:real_scene}.

In this experiment, no bit errors were observed in the 6400-bit transmissions for any of the cameras, indicating that averaging across all pixels helps mitigate noise effects and ensures stable communication.  

Additionally, zoomed-in areas reveal that while compression and differential representation vary between cameras, all successfully recovered the signal. Notably, on the Pixel 8, not all pixels exhibited positive fluctuations, yet the overall average variation across the frame remained positive, enabling correct signal recovery.

\vspace{0.5em}
\noindent{\bf Method Robustness in Diverse Environments.}
Since practical applications involve diverse lighting conditions and scene compositions, additional tests were conducted to evaluate the proposed method's robustness.

First, we used two smartphones to capture 128 bits of data in each of ten scenes to assess accuracy across different scenes. Examples of these scenes are shown in Fig.~\ref{fig:multi_scenes}. This experiment evaluated the impact of scene diversity on detection accuracy and explored the range of environments where the proposed method could be effectively applied.

Additionally, to test robustness against variations in lighting conditions and camera angles, we adjusted the distance between the light source and the object to 88.9, 76.2, and 63.5\,cm (=35, 30, and 25\,in), positioning the smartphones at ten viewing angles for each distance. In each configuration, 128 bits of data were transmitted, allowing us to assess the impact of lighting and camera angles on communication accuracy.

No bit errors were observed under any tested conditions, indicating stable communication performance across diverse scenes and lighting setups.

Using dynamic thresholding, the proposed method demonstrated reliable signal recovery even in scenes with varying reflectance distributions and viewing angles. As shown in Fig.~\ref{fig:multi_scenes}, the recovered signal scales differently across scenes due to reflectance variations, even after normalizing pixel values. Dynamic thresholding ensures consistent signal recovery despite these scale differences, enabling accurate communication in diverse conditions.

\section{Conclusion} 
We have presented the first method to leverage spectral modulation from narrowband LEDs for metadata embedding in video content---imperceptible to human vision yet detectable by RGB cameras. Our approach optimizes spectral modulation to emulate D65 illumination while accounting for both camera and human spectral sensitivities. This enables real-time metadata embedding using ambient lighting without requiring camera modifications. Verification experiments confirmed that these modulations remain undetectable to humans and are robust across different cameras and environmental conditions.

Our findings suggest that spectral modulation from LED lighting holds significant potential for privacy protection and verification applications. By leveraging imperceptible spectral variations rather than intensity differences, our method enables non-intrusive, optically embedded watermarks that preserve the visual experience for human viewers while ensuring reliable detection by camera sensors.

\vspace{0.5em}
\noindent{\bf Limitations and Future Directions.}
While our analysis focuses on static objects, the same conclusion typically holds under small motions when the global image sum is stable, since \(\sum(I_{t-1}-I_{t})=\sum I_{t+1}-\sum I_{t}\). As shown in the supplementary material, deviations arise when \(\sum I_{t}\) changes---e.g., due to non-Lambertian/specular effects, ambient-light shifts, exposure/white-balance adjustments, or occlusions/parallax---which we leave for future evaluation and algorithmic refinements.

Another limitation of our approach is its relatively low communication speed, capped at 15 bps, which may limit data transfer in time-sensitive applications. One potential solution is to increase the data rate by parallelizing communication across multiple wavelength channels, enabling independent data transmission on each channel.

Beyond metadata embedding, imperceptible spectral modulation may have broader implications in optical sensing. While its potential is yet to be fully explored, it could contribute to areas such as hyperspectral imaging and 3D sensing, where spectral control plays a crucial role.

{
    \small
    \bibliographystyle{ieeenat_fullname}
    \bibliography{main}
}


\end{document}